%% file: emnlp2020.tex
\documentclass[11pt,a4paper]{article}
\usepackage{graphicx}
\usepackage{float}
\usepackage{xcolor}
\usepackage{subfigure}
\usepackage{emnlp2021}
\usepackage{times}
\usepackage{latexsym}
\usepackage[english]{babel}
\usepackage{blindtext}
\usepackage{multirow}
\usepackage{siunitx}
\usepackage{booktabs}

\usepackage{tikz}
\usepackage{tikz-qtree}
\usetikzlibrary{trees,calc,arrows.meta,positioning,bending}

\tikzset{
    edge from parent/.style={draw, gray},
    coln/.style={scale=0.6,inner sep=2pt, outer sep=0mm,draw=none,fill=#1, text=white},
    >=latex,
}

\usepackage{microtype}
\usepackage{amsmath}

\newcommand{\avsum}{\mathop{\mathpalette\avsuminner\relax}\displaylimits}
\makeatletter
\newcommand\avsuminner[2]{%
  {\sbox0{$\m@th#1\sum$}%
   \vphantom{\usebox0}%
   \ooalign{%
     \hidewidth
     \smash{\vrule height\dimexpr\ht0+1pt\relax depth\dimexpr\dp0+1pt\relax}%
     \hidewidth\cr
     $\m@th#1\sum$\cr
   }%
  }%
}
\makeatother
\title{Sometimes We Want Translationese} % placeholder title

\date{}
\author{Prasanna Parthasarathi\textsuperscript{1,2},
  Koustuv Sinha\textsuperscript{1,2,3},
  Joelle Pineau\textsuperscript{1,2,3} and
  Adina Williams\textsuperscript{3} \\
  \textsuperscript{1} School of Computer Science, McGill University, Canada \\
  \textsuperscript{2} Quebec AI Institute (Mila), Canada \\
  \textsuperscript{3} Facebook AI Research (FAIR)\\
  \{prasanna.parthasarathi, koustuv.sinha, jpineau, adinawilliams\}\\@\{mail.mcgill.ca, mail.mcgill.ca, cs.mcgill.ca, fb.com\}}

\begin{document}
\maketitle

\begin{abstract}

Rapid progress in Neural Machine Translation (NMT) systems over the last few years has been driven primarily towards improving translation quality, and as a secondary focus, improved robustness to input perturbations (e.g. spelling and grammatical mistakes).  While performance and robustness are important objectives, by over-focusing on these, we risk overlooking other important properties.   In this paper, we draw attention to the fact that for some applications, faithfulness to the original (input) text is important to preserve, even if it means introducing unusual language patterns in the (output) translation.
We propose a simple, novel way to quantify whether an NMT system exhibits robustness and faithfulness, focusing on the case of word-order perturbations.  We explore a suite of functions to perturb the word order of source sentences without deleting or injecting tokens, and measure the effects on the target side in terms of both robustness and faithfulness.  Across several experimental conditions, we observe a strong tendency towards robustness rather than faithfulness.  These results allow us to better understand the trade-off between faithfulness and robustness in NMT, and opens up the possibility of developing systems where users have more autonomy and control in selecting which property is best suited for their use case.
\end{abstract}

% what the perturbations. potentially ranking them with some scores.
% b1 b2 -- different language-pairs and different models
% 

\input{emnlp2020-templates/Sections/Introduction}

\input{emnlp2020-templates/Sections/RelatedWork}

\input{emnlp2020-templates/Sections/Metric}

\input{emnlp2020-templates/Sections/Perturbations}

\begin{figure*}[h!t]
    \setcounter{subfigure}{0}
    \centering
    \subfigure[EN$\rightarrow$DE]{
        \includegraphics[width=0.3\textwidth]{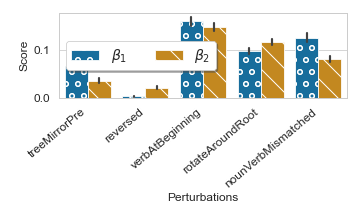}
        }
    \subfigure[EN$\rightarrow$FR]{
        \includegraphics[width=0.3\textwidth]{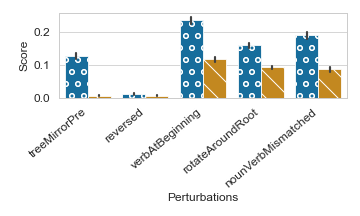}
        }
    \subfigure[EN$\rightarrow$RU]{
        \includegraphics[width=0.3\textwidth]{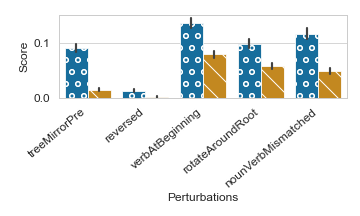}
        }
        
    \subfigure[EN$\rightarrow$JA]{
        \includegraphics[width=0.3\textwidth]{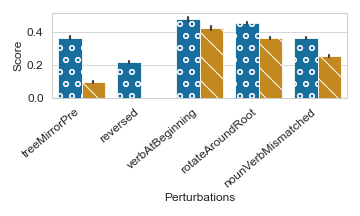}
        }
    \subfigure[EN$\rightarrow$IT]{
        \includegraphics[width=0.3\textwidth]{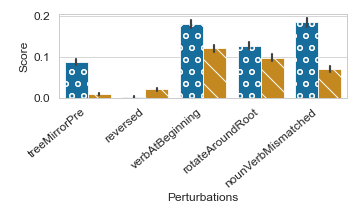}
        }
    % \subfigure[EN-ES]{
    %     \includegraphics[width=0.3\textwidth]{emnlp2020-templates/graphs/partial_levenshtein_beta/Partial_es_levenshtein.png}
    %     }
    \subfigure[EN$\rightarrow$ZH]{
        \includegraphics[width=0.3\textwidth]{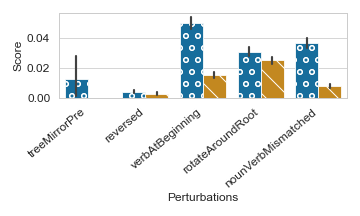}
        }
    \caption{We present the results only from a small set of perturbations to showcase the trend of $\beta_1$ scores being generally higher than $\beta_2$ scores across the different perturbations in different languages.} %This indicates that the translation systems might have a preference to stay robust by ``auto-correcting" the perturbed examples while largely unable to produce faithful translations. The complete result can be found in Appendix.}
    \label{fig:all-perturb-lang-partial}
    
    \subfigure[$\beta_1$]{
        \includegraphics[width=0.45\textwidth]{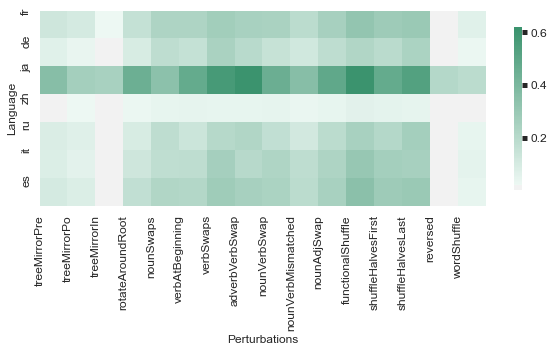}
        }
    \subfigure[$\beta_2$]{
        \includegraphics[width=0.45\textwidth]{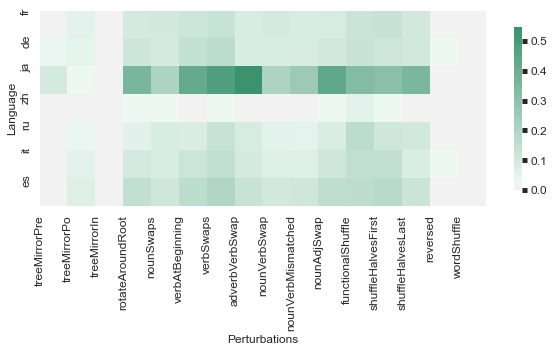}
        }
    \caption{The distribution of average $\beta_1$ and $\beta_2$ across the different target languages show two things. (1) The tendency to be robust in translation for the different languages is indicated by the generally darker heatmap in (a) as compared to (b), (2) within the (a) the PoS based perturbations permit robustness more so than the Tree based and Random indicating that the translation systems find it easier to fix the sentence when majority of the semantics is preserved. (3) The translation system find it difficult to generate faithful translations with most of the perturbations but it was surprisingly low even for the tree based perturbations that skews the semantics.}
    \label{fig:color-map}
\end{figure*}

\input{emnlp2020-templates/Sections/Experiments}

\input{emnlp2020-templates/Sections/Discussion}
\input{emnlp2020-templates/Sections/Conclusion}

\input{emnlp2020-templates/Sections/Ack}
%\section{Acknowledgements}

\bibliography{emnlp2020, anthology}
\bibliographystyle{acl_natbib}

\appendix

\input{emnlp2020-templates/Sections/Appendix}
\end{document}

%% file: emnlp2020-templates/Sections/Introduction.tex
\section{Introduction}

Recent advances in Neural Machine Translation (NMT) have resulted in systems that are able to effectively translate across many languages~\citep{fan2020}, and we have already seen many commercial deployments of NMT technology.
%The state-of-the-art in such systems is most often measured and compared by word overlap scores (BLEU).
Yet some studies have also reported that NMT systems can be surprisingly brittle when presented with out-of-domain data \citep{Luong-Manningiwslt15}, or when trained with noisy input data containing small orthographic \citep{sakaguchi2017robsut, belinkovbisk-2018-synthetic, vaibhav-etal-2019-improving, niu-etal-2020-evaluating} or lexical perturbations \citep{cheng-etal-2018-towards}. Uncovering these sorts of errors has lead the research community to develop new NMT models that are more \emph{robust} to noisy inputs, using techniques such as targeted data augmentation \citep{belinkovbisk-2018-synthetic} and adversarial approaches \citep{cheng2020advaug}. Unfortunately an approach that (over-)emphasized robustness can lead to ``hallucinations''---translating source input to an output that is not \emph{faithful} to the source, and sometimes is even factually incorrect \citep{vinyals2015neural,koehn-knowles-2017-six, wiseman-etal-2017-challenges, nie2019simple,kryscinski2019evaluating, maynez-etal-2020-faithfulness,tian2020sticking, gonzalez-etal-2020-type, xiao2021hallucination}. Moreover, such an approach hinges on the key assumption that orthographic, lexical or grammatical variants in the input are \emph{mistakes}, to be \emph{corrected} by the system in translation. This ignores the wealth of applications where it may be preferable for a system to offer more \emph{faithfulness} to the original text.

%In this work we draw attention to the fact that in some cases, it may be desirable to have an NMT system that is more \emph{faithful} to the original text.  In doing so, we can preserve the source's autonomy in expressing their intent, their preference, perhaps even their personal identity.  In this paper we draw attention to the trade-off between robustness and faithfulness.  We provide a simple approach to audit both properties in NMT systems, for the specific cases of word order anomalies.  We apply such an audit over a class of NMT models and show that existing models tend more towards robustness than faithfulness.  

\begin{figure*}
    \centering
    \subfigure[]{
        \includegraphics[width=0.4\textwidth]{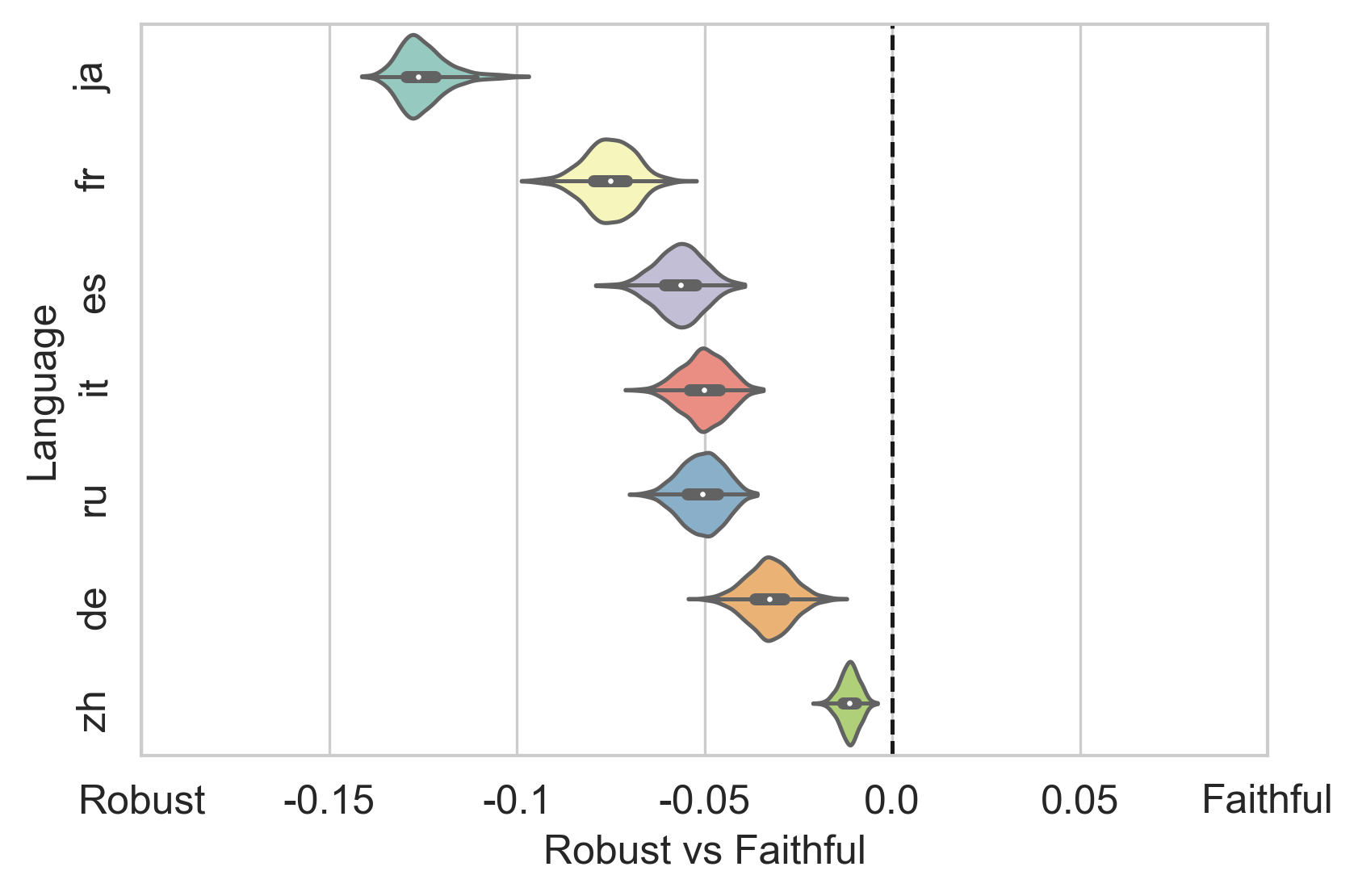}
        }
    \subfigure[]{
        \includegraphics[width=0.5\textwidth]{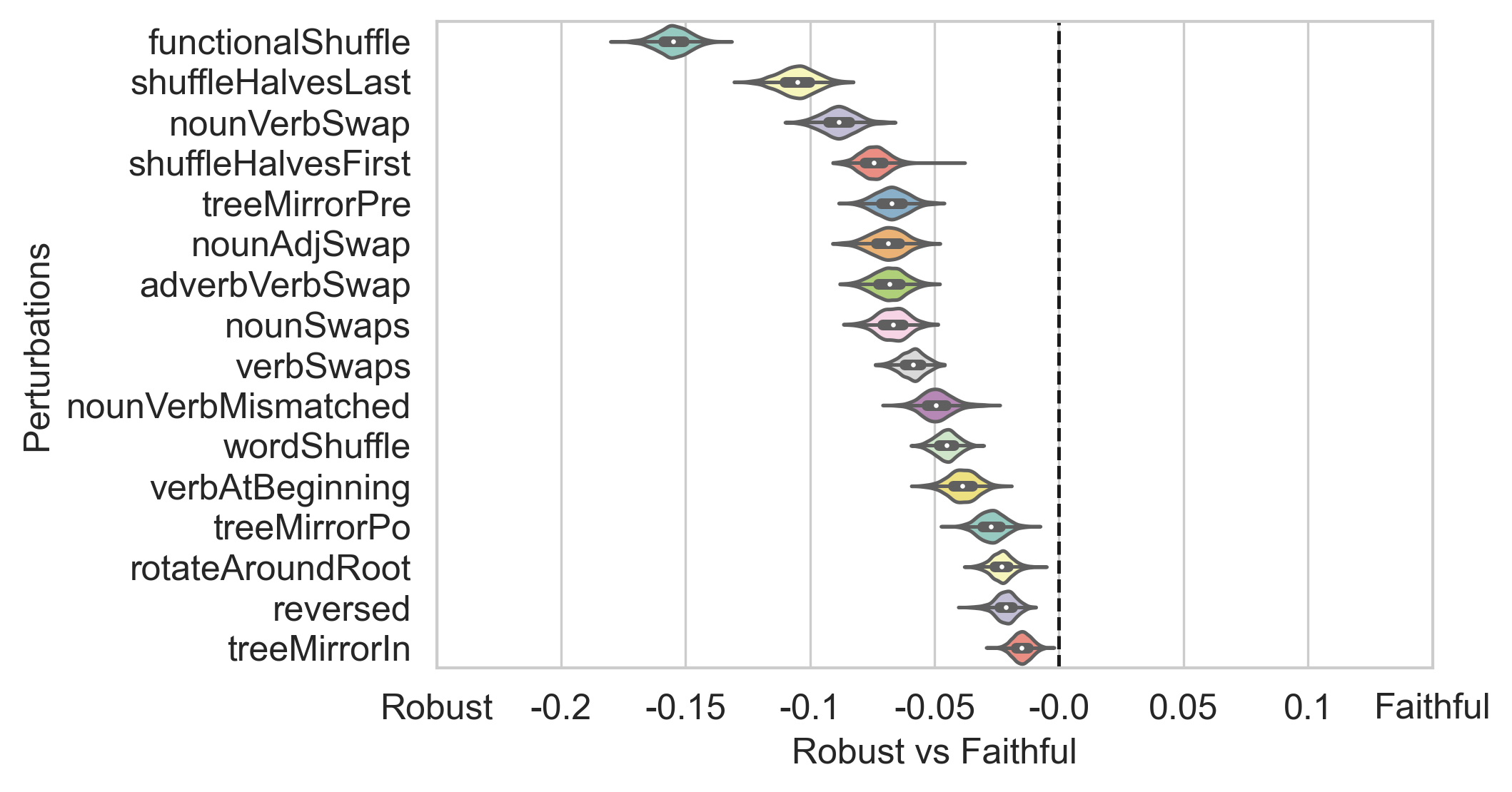}
        }
    \caption{The general trend of the machine translation systems to favor robust or faithful translations is measured by computing the difference between {\bf faithfulness} and {\bf robustness} scores across languages (a) and perturbations (b). The distributions lying close to $-1$ indicate that the translations to that language or the perturbations are more robust than faithful. Since all languages and all permutations fall to the left of 0.0, we can conclude that NMT models are more robust than they are faithful.}
    \label{fig:violin-plot-main}
\end{figure*}

%Given the wealth of research aimed at robustifying NMT systems, it is clear that small spelling perturbations or individual word replacements are a kind of noise that we, as a community, want our NMT systems to be robust to (or, to put it simply, to ignore). However,

%in other cases, one might want a model that will simply output the original target translation when provided with a perturbed source. %, say if someone with an incomplete knowledge of a $2^{nd}$ language wants to translate into a $3^{rd}$. 

It is worthwhile considering the diversity of applications where having a faithful translation is desirable. As a first example, consider an automatic language tutoring system: a (human) second-language learner will often produce language that has grammatical mistakes of various types. This learner can be empowered by having a (AI-produced) faithful translation, so that s/he can see what mistakes were made vs. what would be the more common phrasing. As a second example, recall that many languages, including English, use word order to encode argument structure information (cf. \citealt{isabelle-etal-2017-challenge}), and while most of the time, it may be more logical to see the phrase ``the dog bit the man'', compared to the phrase ``the man bit the dog'', the latter has a very clear meaning that we may wish to preserve in some (albeit rarer) cases.  Third, considering poetry, it is often the case that word order is used to influence rhythm and rhyme.  It would be a shame if all of this music was lost in translation by our state-of-the-art NMT systems.

In short, by their very design, NMT systems tend towards ``normative" language (whether it is spelling, word order or choice of vocabulary).  When we increase their robustness (at least with the solutions proposed to date), we generally enforce even stronger tendency towards the norm, at the expense of diversity of thought, of language and of our very culture.  It is our goal, with this work, to draw attention to this important compromise, and to provide the tools to detect, quantify, and compare such aspects of NMT systems.  We stop short of proposing solutions that can explicitly enable a better trade-off between robustness and faithfulness, and can give the user autonomy and control in specifying their preference in terms of faithfulness.  Though we strongly believe this will be necessary to make real progress in the field.

More specifically, this paper analyzes how existing NMT systems deal with word order permutations in the input.  We investigate 16 unique perturbations that fall into three categories---\emph{Dependency tree based}, \emph{PoS-tag based} and \emph{Random Shuffles}. 
We introduce two novel metrics for evaluating machine translation models' tendency towards robustness and faithfulness. Taking English as the common source, we run a case study with the widely used Transformer-based Helsinki/OPUS machine translation model \cite{opusdataset}, translating into 7 target languages from several families---German, French, Spanish, Russian, Chinese, Japanese, and Italian. 

Across several experimental conditions, we observe a strong tendency towards robustness rather than faithfulness (\autoref{fig:violin-plot-main}).  More specifically, we observe that (1) state-of-the art NMT systems tend towards producing translations that are unaffected by the noisy source (more robust), (2) the accuracy (word overlap BLEU score) correlates with model robustness, (3) certain perturbations involving parts of speech-based word reordering tend to further encourage robustness, and (4) results vary by target language, with the Japanese model producing translations that are most robust but less faithful.  Overall, our analysis suggests that over-focusing on accuracy and robustness may limit richer development and broader usefulness of NMT systems.

%% file: emnlp2020-templates/Sections/RelatedWork.tex
\section{Related Work}

The idea to randomly shuffle linguistic elements to evaluate NLP model performance goes back fairly far \citep{barzilay-lee-2004-catching,barzilay2008modeling}, and has even been used to determine which tasks are ``synax-light'' in human sentence processing \citep{gauthier2019linking}. Recent work on classification tasks, such those on the GLUE benchmark \cite{wang2018glue}, has shown that pre-trained Transformer-based models trained with a masked language modeling objective are shockingly insensitive to word order permutations. \cite{si2019does,sinha2020unnatural, pham2020out, gupta2021bert}.  Given these recent findings, we might expect marked insensitivity to word order permutation in the sphere of generation as well, leading to robust machine translations. 

The mismatching of default word orders between target and source has long been an important consideration for multilingual tasks including automatic machine translation. \newcite{ahmad-etal-2019-difficulties} find that word order agnostic models (recurrent neural networks) trained to  dependency parse can transfer better than word order sensitive ones (self-attention) to distantly related languages. Also in the context of transfer, \newcite{zhao-etal-2020-limitations} propose for reference-free MT that the delta between originally ordered and permuted sentences be used as an evaluation technique. Some works have even aimed to split end-to-end ML into a two step unsupervised process, by first outputting ``translationese'' and then updating the representation to result in a fluent output \citep{pourdamghani-etal-2019-translating}. Even when considering multilingual sequence labeling tasks in general, \newcite{liu-etal-2020-importance, kulshreshtha-etal-2020-cross} find that limiting word order information in the multilingual setting can enable models to achieve better zero-shot cross-lingual performance. Taken together, these works also suggest that our models tend to overfit on source word order to the detriment of that of the target, which might lead one to predict that our models will be more robust than they are faithful in our case as well.

However, NMT systems have use-cases in diverse applications that require the preservation of word order, local syntax and other linguistic components \cite{zhang2020similarity}. Translation systems that are contingent on preserving syntax and semantics are used as interpretors to decode the interaction between components of a neural network \cite{andreas2017translating}. Further, in practical applications like translating a sentence that is a mixture of two different languages requires the MT systems to strike some balance between preserving L1 syntax and/or word-order and correctly adhering to the grammatical rules of L2 \cite{renduchintala-etal-2016-creating}. 

One could think of permuting word order as a way of inducing ``noise'' into the input. Handling samples with noise during inference has been a topic of study that is garnering interest across the research areas in machine learning.
Especially in NLP tasks, where the end-user could be a human, benchmarking the robustness of NLP systems is done by evaluating a model's performance on willfully perturbed examples that could potentially expose fragility of the systems \cite{goodfellow2014explaining, fadaee2020unreasonable}. Towards averting such scenarios, efforts along the lines of building robust models with Adversarial training has been a common topic of study in natural language processing \cite{adversarialnlg, wu2018adversarial}.

Our word order perturbation also share some points of synergy with work across NLP that aims to devise supplementary heuristics to explicate the inner workings of our machine learning systems. For specific NLP tasks, probe tasks are engineered to measure specific kinds of linguistic knowledge encoded in the systems \cite{conneau1germa, sheng-etal-2019-woman, kim2019probing,  jeretic-etal-2020-natural,   parthasarathi2020evaluate,  ribeiro-etal-2020-beyond}. Swapping the arguments of verbs is a classic way to measure the effects of word order both in humans \citep{frankland2015architecture, snell-grainger-2017-sentence} and in models, largely because changing the order of verbal arguments maintains high word overlap between related examples \citep{wang2018glue, kann-etal-2019-verb, mccoy-etal-2019-right}; However, although limited word order permutation is applied in this case, it is generally restricted to licit, grammatical sequences of words. When perturbation has been used to evaluate model performance, the utilized perturbation functions have been predominantly fairly simple, including reverse and word shuffle, and usually target only single sentences \cite{ettinger2020bert, li2020contextualized, sinha2020unnatural}. For tasks like dialogue prediction that requires multiple input sentences, perturbation functions like reordering the conversation history are adopted \cite{sankar2019neural}. To the best of our knowledge, the proposed set of perturbation functions is the most detailed set explored thus far, perturbing not only tokens, but PoS and dependency structure.

Changing the order of words in the context of NMT also its roots in classical, syntax based models that used parses (of various kinds) to pre-order abstract syntactic representations as an early step in a multi-step translation pipeline from source to target (\citealt{collins-etal-2005-clause, khalilov-etal-2009-coupling, dyer-resnik-2010-context, genzel-2010-automatically, khalilov-simaan-2010-source, miceli-barone-attardi-2013-pre} i.a.). Our approach differs from these approaches in that our main aim is not to incorporate word order changes into the translation pipeline itself, but is instead to use them to better understand the behavior of NMT models.

%Finally, conditional generation of language with neural language models have documented hallucinative behavior \cite{nie2019simple,xiao2021hallucination}. \citet{kryscinski2019evaluating, maynez2020faithfulness} observe that text summarization models may be less faithful to the input than being hallucinative. While hallucinations of non-existent entities could potentially decrease the credibility of the summarization systems, in this work, we pose a neutral stand on hallucinating unperturbed sentences in translations. Such cautious hallucinations improve the robustness of the translation systems to noise at the same time staying faithful to the translations have a different set of downstream tasks \cite{edunov2018understanding}.

%% file: emnlp2020-templates/Sections/Metric.tex
\section{Metrics}

In this section, we define several metrics to determine whether current SOTA NMT models preferentially output robust or faithful translations in the face of word order permuted source sentences.

Let $g_e$ be an English ($E$) sentence from the test set of a translation task ($E \rightarrow O$), where the output language $O\ \sim$ {
\emph{German} (de), \emph{French} (fr), \emph{Spanish} (es), \emph{Italian} (it), \emph{Russian} (ru), \emph{Japanese} (ja), \emph{Chinese} (zh)}. Let  $\Psi$ denote a perturbation function such that $g_e^{-} \gets \Psi\left(g_e\right)$. Let $\Phi_{E\rightarrow O}$ denote a   translation pipeline from English ($E$) to another language ($O$),

\begin{equation}
    g_{o} \gets \Phi_{E\rightarrow O}\left(g_e\right)
\end{equation}
\begin{equation}
    g^{-}_{o} \gets \Phi_{E\rightarrow O}\left(g^{-}_e\right)
\end{equation}

Let $\kappa\left(s_i,s_j\right)$ be a scoring function that rates the similarity between two sentences($s_i$ and $s_j$), where $s_i,s_j \in L_*$. The choice of $\kappa$ can be any of the widely used sentence similarity metrics like BLEU \cite{papineni-etal-2002-bleu}, METEOR \cite{lavie2007meteor}, ROUGE \cite{lin2004rouge}, or Levenshtein-distance \cite{levenshtein1966binary}. For our purposes, we will use BLEU-4, and Levenshtein score as choices of $\kappa$ denoted by a $B$ or $L$ in the superscript respectively. The value of $\kappa$ linearly scales with the similarity between $s_i$ and $s_j$.

We define three metrics $\beta_1$, $\beta_2$ and $\alpha$ in Equations \ref{eqn:beta_1}, \ref{eqn:beta_2} and \ref{eqn:alpha} respectively. $\beta_1$ measures \textbf{robustness} to perturbation by scoring the similarity between the translation of a perturbed sentence in source and the gold sentence in target:

\begin{equation}
    \beta_1 \gets \avsum_{g_e}\kappa\left(g_o,\Phi_{E\rightarrow O}\left(g_o^-\right)\right)
    \label{eqn:beta_1}
\end{equation}

where $\avsum_{g_e}$ denotes $\frac{1}{N_{E\rightarrow O}^{\Psi}}\sum_{g_e} (\cdot)$; $N_{E\rightarrow O}^{\Psi}$ denote the cardinality of the set of samples perturbed by $\Psi$.

%We should intuitively expect high $\beta_1$ when the translation of the perturbed input and the gold target sentence are similar, meaning that the model has managed to surface a translation that is very similar to the gold.

$\beta_2$ is computed as a similarity score between the translation perturbed source sentence and applying the same perturbation operation on the target sentence to measure degree of \textbf{faithfulness} of translations by machine translation system: 

\begin{equation}
    \beta_2 \gets \avsum_{g_e} \kappa\left(g_o^-,\Phi_{E\rightarrow O}\left(g_e^-\right)\right)
    \label{eqn:beta_2}
\end{equation}

The \textbf{difficulty} of the perturbation function is measured with $\alpha$, which scores the similarity between perturbed sentence and the unperturbed sentence in the source language. 

\begin{equation}
    \alpha_e \gets \avsum_{g_e} \kappa\left(g_e,g_e^-\right)
    \label{eqn:alpha}
\end{equation}

$\beta$ measures the standard translation performance metric on any given source-target sentence pair.

\begin{equation}
    \beta \gets \avsum_{g_e} \kappa\left(g_o,\Phi_{E\rightarrow O}\left(g_e\right)\right)
    \label{eqn:beta}
\end{equation}

%% file: emnlp2020-templates/Sections/Perturbations.tex
\section{Perturbations}\label{sec:perturbations}

\begin{figure*}[ht]
    \centering
    \subfigure[]{
    \fbox{\begin{minipage}{0.45\textwidth}
    \small
    \textbf{TreeMirrorPost:} to live a decent place he could n't find Tom said .\\

    \textbf{TreeMirrorPre:} said find place live to a decent he could n't Tom .\\

    \textbf{TreeMirrorIn:} live to place a decent find he could n't said Tom .\\
    
    \textbf{RotateAroundRoot:} live find said Tom he could n't a decent place to .\\
    
    \textbf{WordShuffle:} place to could live said decent a Tom n't find he .\\
    
    \textbf{Reversed:} live to place decent a find n't could he said Tom .\\
    \end{minipage}
    \label{fig:perturb-examples-a}
    }
    }
    \subfigure[]{\fbox{\begin{minipage}{0.45\textwidth}
    \small
    \textbf{VerbSwaps:} Tom live he find n't said a decent place to could .\\

    \textbf{NounSwaps:} Tom said a decent place could n't find he to live .\\

    \textbf{NounVerbSwaps:} said Tom could he n't a decent place find to live .\\
    
    \textbf{NounVerbMismatched:} live a decent place find could n't he said to Tom .\\

    \textbf{ShuffleFirst:} he Tom find could said n't a decent place to live .\\

    \textbf{ShuffleLast:} Tom said he could n't find a decent live place to .\\
    \end{minipage}
    }\label{fig:perturb-examples-b}
    }
    \caption{Effect of the different perturbation functions on the sentence --- \emph{Tom said he could n’t find a decent place to live.} The perturbation functions do not inject new tokens or delete a token to perturb the sentence.}
    \label{fig:perturb-examples}
\end{figure*}

We propose $16$ different functions to perturb the structure of an input sentence. The perturbations can be broadly classified in three categories---\emph{Random Shuffles}, \emph{PoS-tag Based} and \emph{Dependency Tree Based}---comprised of 4, 8, and 4 perturbation functions respectively. The functions vary in complexity and linguistic sophistication so that twe can score whether a model performs faithful translations or stays robust to the perturbed inputs. We applied all perturbations in seven languages---\emph{de}, \emph{fr}, \emph{ja}, \emph{ru}, \emph{zh}, \emph{it}, and \emph{es}---and describe each perturbation in turn below. See \autoref{fig:perturb-examples} for a selection of examples, and \autoref{tab:sampledist} in the Appendix for more information on counts of perturbations by language.

Some perturbations we explore are ``possible'', in the sense that applying them will result (in most cases) in a grammatical sentence (either in the source language, or in some version of another existing language that is instead supplied with the words of the source). Others are ``impossible'' \citep{moro2015boundaries, moro2016impossible}. For example, it has been long noticed that human grammar rules operate on hierarchical structure resulting in rules of the form ``move the hierarchically closest auxiliary when forming a question'' as opposed to ``move the linearly closest auxiliary in question formation''. Standard American English exemplifies this: when we for a question from the sentence ``The man who is tall was happy'', we say ``Was the man who is tall happy?'' as opposed to ``Is the man who tall was happy?'' (\citealt{mccoy-etal-2020-syntax}, cf. \citealt[Ch. 3]{chomsky1957syntactic}). To explore more fully the behavior of the NMT models, we include several permutations that do not adhere to the descriptive rules of the source language , let alone any human grammars across all known languages (i.e., are ``impossible''). 

\subsection{Random Shuffles}
The perturbations in the Random bin treat the sentence as though it were a mere sequence of tokens. They merely reorder the tokens without any reference to their higher order linguistic properties (i.e., PoS or dependency information). Thus, random perturbations can be seen as the most basic type of ``impossible'' word order perturbation. We use three different random shuffles--- \emph{Word-Shuffle}, \emph{Shuffle-First-Half}, \emph{Shuffle-Last-Half} and \emph{Reversed}---none of which result in any recognizable linguistic structure. \emph{Word-Shuffle} shuffles the entire sentence at random (cf. \citealt{sinha2020unnatural}); for a sentence of length $n$, there are $(n-1)!$, possible random permutations. \emph{Shuffle-First-} and \emph{Shuffle-Last-Halves} shuffle only the corresponding half of a sentence while keeping the other half unperturbed. 
\emph{Reversed} reverses the token ordering in a sentence.

\begin{figure}[ht]
    \centering
    \subfigure[Original]{
    \resizebox{0.45\columnwidth}{!}{
    \begin{tikzpicture}[
    level/.style={level distance=8mm},
    level 1/.style={sibling distance=12mm},
    level 3/.style={sibling distance=10mm},
    block/.style={
      rectangle,
      draw=blue,
      thick,
      fill=blue!20,
      align=center,
      rounded corners,
    }
    ]

\node (root) {ROOT}
    child {node {VERB}
        child {node {NSUBJ}
            child {node {PROPN}
                child {node[text=red]{Joe}
                }
            }
        }
        child {node[text=red]{waited}
        }
        child {node {NOUN}
            child {node{ADP}
                    child {node[text=red]{for}
                    }
                }
            child {node{DET}
                child {node[text=red]{the}
                    }
                }
            child {node[text=red]{train}
                }
        }
};

% \draw[thick, blue!60!black] (root.east)
%     edge[out=-10,in=45,->,looseness=1.4] node [midway,right,xshift=.3em,coln=blue!55] {Root}
%         ($(eat.north east)+(-1mm,-1mm)$);
% \draw[thick, red] (eat.west)
%     edge[out=180,in=270,->] node [midway,left,xshift=-.3em,coln=red!60] {NEG}
%         (not.south);

\end{tikzpicture}
}
}
\subfigure[Mirrored]{
    \resizebox{0.45\columnwidth}{!}{\begin{tikzpicture}[
    level/.style={level distance=8mm},
    level 1/.style={sibling distance=12mm},
    level 3/.style={sibling distance=10mm},
    block/.style={
      rectangle,
      draw=blue,
      thick,
      fill=blue!20,
      text width=5em,
      align=center,
      rounded corners,
      minimum height=2em
    }
    ]

\node (root) {ROOT}
    child {node {VERB}
        child {node {NOUN}
            child {node[text=red]{train}
                }
            child {node{DET}
                child {node[text=red]{the}
                    }
                }
            child {node{ADP}
                    child {node[text=red]{for}
                    }
                }
        }
        child {node[text=red]{waited}
        }
        child {node {NSUBJ}
            child {node {PROPN}
                child {node[text=red]{Joe}
                }
            }
        }
};

% \draw[thick, blue!60!black] (root.east)
%     edge[out=-10,in=45,->,looseness=1.4] node [midway,right,xshift=.3em,coln=blue!55] {Root}
%         ($(eat.north east)+(-1mm,-1mm)$);
% \draw[thick, red] (eat.west)
%     edge[out=180,in=270,->] node [midway,left,xshift=-.3em,coln=red!60] {NEG}
%         (not.south);

\end{tikzpicture}
}
}
    \caption{With the root at the top, an InOrder traversal (Left-Root-Right) of the tree returns the exact sentence. In the tree based perturbations, we mirror the dependency tree and perform InOrder, PreOrder (Root-Left-Right) and PostOrder (Left-Right-Root) to perturb the sentence.}
    \label{fig:dependency-tree}
\end{figure}

\subsection{Part-of-Speech tag Based Perturbations}
This set of perturbations uses the PoS tag information from a parser to generate perturbations for a sentence, so that we can localize any effects of robustness or faithfulness to particular linguistic categories. %Our Part-of-Speech tag based perturbations include two types: perturbations that are either more ``possible'' (including all swaps that swap two elements with the same POS) and ``impossible'' ones (which swap tokens with different PoS, or ma. 

\paragraph{PoS Swaps.} When a sentence has more than one token with a particular PoS, the positions of those tokens are exchanged without affecting the rest of the sentence structure.\footnote{Modulo cases where person agreement might be affected, for example when verb-swapping ``am'' for ``are'' in \textit{I am happy that they are here}$\rightarrow$ \textit{I are happy that they am here.}} Although the meanings of the sentences are altered, the result generally is grammatical (or near grammatical, see \autoref{fig:perturb-examples}(b)), meaning that these swaps are ``possible''. In this class of permutations, we consider Noun swaps and Verb swaps. 

\paragraph{PoS$_X$-PoS$_Y$ Swaps.} The position of a token with a particular PoS tag $X \in {noun, adv}$ is interchanged with the linearly closest token with PoS tag $Y \in {verb, adj}$ leaving the rest of the sentence unperturbed. In this class, we consider Adverb-Verb swaps and Noun-Adjective swaps (which tend to result in grammatical sentences), as well as Noun-Verb swaps (which tend to result in ungrammatical sentences).

\paragraph{PoS\textsubscript{Noun}-PoS\textsubscript{Verb} Mismatched Swaps.} While Noun-Verb Swap replaces each noun with the verb closest to it, the mismatched swap exchanges the position of a noun with the verb \emph{farthest} from it, which results in displacing all verbs and nouns from their original positions. 

\paragraph{Functional Shuffle.} Functional tokens--conjunctions, propositions and determiners---are shuffled and rearranged in positions of a different functional token to generate the perturbed sentence. 

\paragraph{Verb-At-Beginning.} This perturbation moves a verb to the beginning of the sentence as a prefix without disturbing the remaining relative positions within the text.

\subsection{Dependency Tree Based}
The dependency tree structure of a sentence conveys its grammatical structure. When one perturbs the dependency tree in a language like English---which expresses verb-argument relationships largely via word order---there could be several effects: the semantics of the sentence will be changed, and the base word order might now be indicative of a different family of languages. Therefore, we investigate dependency tree perturbations with an eye towards determining whether perturbations that result in sentence structures from another family (e.g., Japanese) will result in more faithful translations.  

\paragraph{Tree Mirror (Pre/Post/In).} While an In-Order traversal of a sentence's dependency tree (\autoref{fig:dependency-tree}) provides the right parse of the sentence, we perform Pre-Order, Post-Order and In-Order traversals on the mirrored dependency tree. Although the perturbed sentences largely preserve each word's position with respect to its local neighbors, since they are ungrammatical, their meanings (if there are any) are much harder to immediately appreciate. 
\paragraph{Rotate Around Root.} The sentence is perturbed by rotating the tree around its root and then subsequently performing an In-Order traversal.

\subsection{Distribution}
%\label{sec:bins-beta-comparison}
To better understand the results of our perturbations on the downstream translations, we first measure the mutual similarity between the perturbation functions shown in \autoref{fig:bins-beta-and-alpha-comparison-1} measured as $\kappa\left(\Psi_i(s), \Psi_j(s)\right)$ for every perturbation function $i$ and $j$ highlight the differences between the three categories of perturbations.

We observe that the dependency tree-based perturbation functions have less overlap with the PoS tag-based perturbations across languages, but higher intra-category similarity scores.  Similarly the PoS tag-based functions have understandably higher similarity with other PoS tag-based functions than with Shuffle or Dependency tree perturbation functions.

% \begin{table}[htbp]
%     \scriptsize
%     \begin{tabular}{|p{1.6cm}|p{0.5cm}|p{0.5cm}|p{0.5cm}|p{0.5cm}|p{0.5cm}|}
%     \hline
%     \textbf{Lang}/\textbf{Bins} & \textbf{1} & \textbf{2} & \textbf{3} & \textbf{4} & \textbf{5} \\
%     \hline
%       \textbf{de} & $0.85$  & $0.016$ & $0.045$ & $0.032$ & $0.055$ \\
%       \textbf{fr} & $0.82$ & $0.040$ & $0.058$ & $0.036$ & $0.039$ \\
%       \textbf{ru} &  $0.89$ & $0.01$ & $0.03$ & $0.02$ & $0.04$ \\
%       \textbf{ja} & $0.35$ & $0.16$ & $0.21$ & $0.13$ & $0.12$ \\
%       \textbf{es} & $0.80$ & $0.02$ & $0.06$ & $0.04$ & $0.06$ \\
%       \textbf{it} & $0.86$ & $0.012$ & $0.035$ & $0.029$ & $0.052$  \\
%       \textbf{zh} & $0.95$ & $0.001$ & $0.010$ & $0.009$ & $0.025$ \\
%     \hline
%     \end{tabular}
%     \caption{BLEU-4 metric}
%     \label{tab:distribution}
% \end{table}

%% file: emnlp2020-templates/Sections/Experiments.tex
\section{Experiments}

We experiment with the state-of-the-art Transformer translation models %in \emph{Helsinki/OPUS} 
over 7 different languages paired with English---French (fr), German (de), Russian (ru), Japanese (ja), Chinese (zh), Spanish (es), and Italian (it). None of our methods are architecture dependent; We focus on the OPUS translation models \citep{opusdataset} which are based on a standard transformer setup with 6 self-attentive layers in both encoder and decoder networks, with 8 attention heads per layer. Our experiments have a twofold objective: (1) compute the robustness ($\beta_1$) and faithfulness ($\beta_2$) of the translations in different languages when the input is perturbed, and (2) analyse the $\beta_1$ and $\beta_2$ scores with different levels of perturbations. 

\section{Results}

\subsection{Faithfulness vs. Robustness}

For each language paired with English, we perturb the gold sentences both in English and in the target language using the perturbation functions proposed in \S\ref{sec:perturbations}. We measure $\beta^B_1$ and $\beta^B_2$  scores (with BLEU-4) for every perturbation function (see \autoref{fig:all-perturb-lang-partial}\footnote{The complete list is in Appendix \S \ref{sec:complete-table}}).  

We observe that $\beta_1$ scores are generally higher than $\beta_2$ scores for all the perturbation functions across all the languages, indicating that the OPUS translation system is largely unphased when presented with  \emph{unnatural}, \emph{ungrammatical} input (see \autoref{fig:violin-plot-main}). Given these results, the model acts as though it makes an intermediate ``hallucination'' that somehow either recreates the unperturbed input before then translating it, or ``hallucinates'' an unperturbed target without much reference to the perturbed source.

In \autoref{fig:color-map}, models seem to ignore the precise word order they are presented with: Compare the heat maps showing higher $\beta_1$ than $\beta_2$ values on average across languages. Across the different languages, models tend to recover more when faced with \emph{PoS tag-based} perturbations: Figure \ref{fig:color-map}(a) generally shows darker shades for PoS tag-based perturbations than for the others. This means that models find it harder to ignore word order for sentences perturbed with \emph{Dependency tree-based} and \emph{Random} perturbations than with \emph{PoS tag-based} ones.

% \begin{figure*}[h]
%     \centering
%     \subfigure[Bin 1]{
%          \includegraphics[width=0.45\textwidth]{emnlp2020-templates/graphs/Violin/binned_beta1_and_beta_2_Bin 1.png}
%          }
%     \subfigure[Bin 2]{
%          \includegraphics[width=0.45\textwidth]{emnlp2020-templates/graphs/Violin/binned_beta1_and_beta_2_Bin 2.png}
%          }
         
%     \subfigure[Bin 3]{
%          \includegraphics[width=0.45\textwidth]{emnlp2020-templates/graphs/Violin/binned_beta1_and_beta_2_Bin 3.png}
%          }
%     \subfigure[Bin 4]{
%          \includegraphics[width=0.45\textwidth]{emnlp2020-templates/graphs/Violin/binned_beta1_and_beta_2_Bin 4.png}
%          }
%     \caption{We split the samples into bins according to their difficulty and measured $\beta_2^L$ and $\beta_1^L$. As \textit{Bins 1 and 2} contain the easier examples, as well as the majority of the data, we conclude that %The fraction of samples in across the different bins shows that fraction of samples with lower $\beta_1^L$, indicates the models are more robust than faithful. Greater fraction of samples with higher $\beta_2^L$ score in Bin-4 than shows that the model might find it likely easier to be robust than being faithful in translation. Chinese dataset's distribution of $\beta_1$ and $\beta_2$ largely pertaining to Bin 4 indicate the difficulty of the dataset.
%     }
%     \label{fig:samples-across-bins-distribution}
% \end{figure*}

\subsection{Patterns in $\beta_1$ and $\beta_2$}
Given our results, we would like to know whether there are any particular properties of particular examples or of permutations which lead models to be more or less robust. Towards that end, we observe the correlations between (a) $\beta$ vs $\beta_1$/$\beta_2$ (b) $\beta_1$ vs $\beta_2$, and (c) $\beta_1$/$\beta_2$ vs $Length$ of source sentence.

\begin{figure}
    \centering
    \includegraphics[width=\columnwidth]{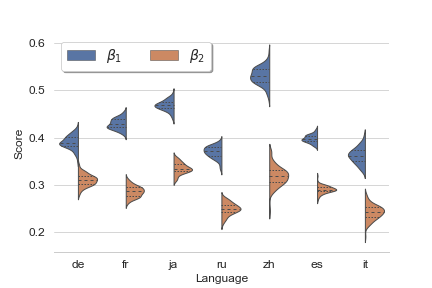}
    \caption{The correlation of $\rho(\beta^B_1,\beta^B)$ as $\beta^B_1$ and $\rho(\beta^B_2,\beta^B)$ as $\beta^B_2$ shows that the robustness of the translation system has a strong correlation with the performance of the machine translation system. The faithful translations have a weak correlation, indicating that the easier to translate examples are difficult for the model to do word-to-word translations on.}
    \label{fig:correlation-betas}
\end{figure}

\paragraph{$\beta$ vs $\beta_1$/$\beta_2$.}
We find that our $\beta_1$ does correlate with BLEU-4 on the translation of the original, unperturbed gold English sentence and gold target language. We compute correlations of $\beta_1$ and $\beta_2$ with $\beta$ in \autoref{fig:correlation-betas}. The Spearman's rank correlation between $\beta_1$ and BLEU is larger than between $\beta_2$ and BLEU; in the former we observe a medium strength effect and in the latter a small effect, although language does play a role (e.g., Chinese has the largest $\beta_1$ correlation with BLEU, but among the smallest $\beta_2$ correlation with BLEU). 

\begin{figure*}
    \setcounter{subfigure}{0}
    \centering
    \subfigure[$\beta_1$ vs $\beta_2$]{
        \includegraphics[width=0.3\textwidth]{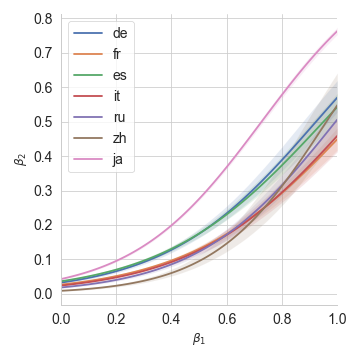}
        }
    \subfigure[$\alpha_e$ vs \emph{Length}]{
        \includegraphics[width=0.3\textwidth]{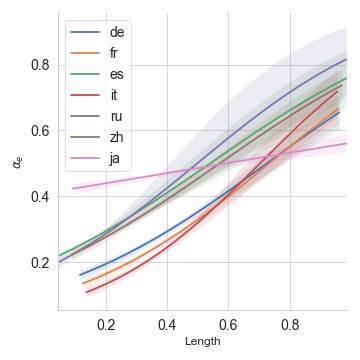}
        }
        
    \subfigure[$\beta_1$ vs \emph{Length}]{
        \includegraphics[width=0.3\textwidth]{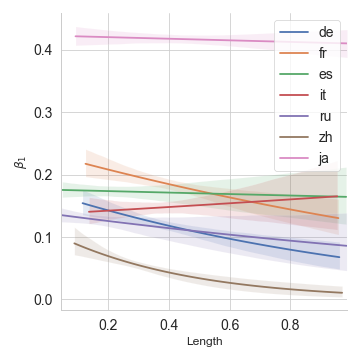}
        }
    \subfigure[$\beta_2$ vs \emph{Length}]{
        \includegraphics[width=0.3\textwidth]{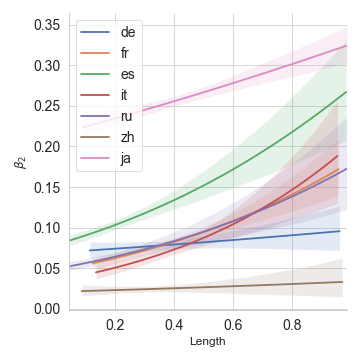}
        }
    
    \caption{We observe the length of the source sentences to differently correlate with the two scores. The robustness score, $\beta_1$, is higher for shorter source sentences, while the opposite is true for $\beta_2$. This suggests that the model's ability to see through the syntactic errors has a limitation on the length. Also, the model being able to stay faithful in longer sentences can be explained with higher $\alpha_e$ longer sentences score. With longer sentences hence being relatively less difficult, the $\beta_2$ scores are higher.}
    \label{fig:beta-scores-length-correlation}
\end{figure*}

\paragraph{$\beta_1$ vs $\beta_2$.} \autoref{fig:beta-scores-length-correlation}(a) shows that the correlation between robustness and faithfulness to be present, but weak. By definition, the model can either be faithful or robust and when it is both, then that suggests only a higher $\alpha_e$ or a lower perturbation difficulty. Usually this occurs when sentences are very short---for short sentences, fewer permutations are possible, and different permutation functions are more likely to collapse onto the same word orders. 

\paragraph{$\beta_1$/$\beta_2$ vs \emph{Length}.} The length of the source sentence has different effects on the scores depending largely on language. But, it is intuitive to understand that the model is better able to fix a word order perturbation when the sentences are short, resulting in higher $\beta_1$ score for shorter sentences. The opposite is true for $\beta_2$ where longer sentences generally have higher $\beta_2$ score. 

% split the examples based on their $\alpha_E^L$ scores, a proxy to the difficulty of the perturbation functions, into 4 different bins. We placed the perturbed examples with the lowest edit-distance with the original source sentence into \emph{Bin 1} while the maximally distant samples were placed in \emph{Bin 4}. Across the bins (\autoref{fig:bins-beta-and-alpha-comparison-2}), we compute (a) $\beta^B_1$ (b) $\beta^B_2$ and (c) translation quality of the unperturbed source as $\beta^B$.

%We observe that the gap between $\beta$ and $\beta_1$ increases with the bin indices, which indicates that robustness decreases when the perturbed input is of higher difficulty. Although the overall model's tendency is produce robust translations, the existence of an increasing gap as examples become more difficult strongly suggests room for improvement.

There is some relationship between which permutation function generated a permuted example and its $\alpha_E$ score (\autoref{fig:alpha-scores}). The top 5 permutation functions with high $\alpha_E$ scores---\{\emph{shuffleHalvesLast}, \emph{shuffleHalvesFirst}, \emph{verbAtBeginning}, \emph{nounVerbSwap}, \emph{nounVerbMismatched}\}---and with low $\alpha_E$ scores---\{\emph{treeMirrorPost}, \emph{wordShuffle}, \emph{reversed}, \emph{treeMirrorIn}, \emph{treeMirrorPre}\}. The mix of examples from different perturbation categories at different levels of high $\alpha_E$ score, as well as the fact that $\beta_1$ scores are higher than $\beta_2$, suggests that models' attempting to self-correct the perturbed input may not be because they understand language, but instead it might be due to correlations between certain n-grams in the sentence. We also observe that $\beta_1$ decreases with increasing $\alpha_E^L$, which also supports this argument.

% \begin{figure*}[h]
%     \setcounter{subfigure}{0}
%     \centering
%     \subfigure[EN-DE]{
%         \includegraphics[width=0.45\textwidth]{emnlp2020-templates/graphs/beta_comparison/de_bleu-4.png}
%         }
%     \subfigure[EN-FR]{
%         \includegraphics[width=0.45\textwidth]{emnlp2020-templates/graphs/beta_comparison/fr_bleu-4.png}
%         }
%     \subfigure[EN-RU]{
%         \includegraphics[width=0.45\textwidth]{emnlp2020-templates/graphs/beta_comparison/ru_bleu-4.png}
%         }
        
%     \subfigure[EN-JA]{
%         \includegraphics[width=0.45\textwidth]{emnlp2020-templates/graphs/beta_comparison/ja_bleu-4.png}
%         }
%     \subfigure[EN-IT]{
%         \includegraphics[width=0.45\textwidth]{emnlp2020-templates/graphs/beta_comparison/it_bleu-4.png}
%         }
%     % \subfigure[EN-ES]{
%     %     \includegraphics[width=0.45\textwidth]{emnlp2020-templates/graphs/beta_comparison/es_bleu-4.png}
%     %     }
%     \subfigure[EN-ZH]{
%         \includegraphics[width=0.45\textwidth]{emnlp2020-templates/graphs/beta_comparison/zh_bleu-4.png}
%         }
%     \caption{BLEU-4 All perturbations}
% \end{figure*}

%% file: emnlp2020-templates/Sections/Discussion.tex
\section{Discussion}
One way to think about the models' tendency towards behaving robustly is to take them to be hallucinating an unperturbed response even when the word order of the original is perturbed. 
% KS: this is not clear at all to me
%The distribution of samples across the bins constructed with $\beta_1^L$ and $\beta_2^L$ in Figure \ref{fig:samples-across-bins-distribution} show that there is a slightly larger fraction of samples with higher similarity measured with $\beta_1^L$ in Bins 1 and 2 than with $\beta_2^L$. 
Observing the difference between $\beta_1$ and $\beta_2$ (\autoref{fig:violin-plot-main}) %\ref{fig:correlation-betas} 
shows a ranking across languages, and with perturbation functions. Amongst the languages analysed, we find that Japanese is generally more robust than the other languages. However, we note that our findings could also be attributed to Japanese sentences being less difficult even after perturbed; the $\alpha_e$ of the Japanese translation model is much higher than for the rest of the languages (\autoref{fig:alpha-scores} in \autoref{alpha-results}). Similarly, the weak $\beta_1$ and $\beta_2$ scores of Chinese translation model could also be attributed to the general poor performance of the translation systems for the language (\autoref{tab:recorded-mt-performances} show that the $\beta$ scores of the Chinese model itself are too low).

Among the perturbation functions, \emph{FunctionalShuffle} evoked the most robust generation across all languages while models were most faithful on \emph{TreeMirrorIn} and \emph{Reversed}. Recall, however, the fact that all languages fall to the left of $0$ in both (a) and (b) in \autoref{fig:violin-plot-main} reflecting that all models are reasonably robust. More work is needed to suggest clear ways of training a model to maximize its faithfulness or robustness. We believe our perturbation methods can be used to guide model selection by helping to determine just how faithful or robust a model should be based on specific downstream requirements. %While there is no clear suggestion to improve the language generation to be more faithful or robust, it is useful to analyse the language generation systems with the proposed metrics and perturbations to appropriately select a model appropriately. 

Similar to the flips in the predictions observed by \citeauthor{sinha2020unnatural}, we observe in a fraction of the translations the $\beta_1$ scores to be greater than $\beta$ (\autoref{tab:flips-NMT} in Appendix). This suggests that the model might require the source sentences to be in a particular order for the expected translations. Although rare, examples for which reordering the source results in a better target translation do exist---our work opens up potential avenues for probing datasets for flips as a way to measure ``unnaturalness'' of models' translation algorithms.

%% file: emnlp2020-templates/Sections/Conclusion.tex
\section{Conclusion}
In this work, we analyze whether neural machine translation (NMT) models produce outputs that are robust or faithful in the face of permuted source sentences. 
It is important to understand how NMT systems behave on such malformed input---should a model be robust and risk ``hallucinating" from the input, or should it be faithful, taking the input at face-value, and provide faithful, word-by-word translations. Particular examples might differ in whether a robust or a strongly faithful approach is warranted (for example, we wouldn't want to badly translate poetry that was using nonstandard word order for creative effect). Our novel metrics and perturbation functions allow one to quantify how systems strike a balance between robustness and faithfulness in NMT, both on the corpus level and at the level of particular examples.
We find that state-of-the-art Transformer based Helsinki/OPUS models
% translating from English into 7 different target languages
have a general tendency to be robust, i.e., to be unperturbed by word-order perturbed inputs. We also find that the robustness of a model has strong correlation with the performance of the translation task---which is an indicator that our training data or model's inductive bias nudges translation output to be more robust.  %We propose a suite of novel metrics and perturbation functions to study the robustness-faithfulness choices of state-of-the-art Transformer based Helsinki/OPUS models. 

%% file: emnlp2020-templates/Sections/Ack.tex
\section*{Acknowledgements}

We thank Dieuwke Hupkes, Douwe Kiela, Robin Jia, Molly FitzMorris, Adi Renduchintala, Mona Diab, Hagen Blix, and Radhika Govindarajan for the suggestions and discussions that helped shape this work. Further, we thank McGill University and ComputeCanada for the computing resources.

%% file: emnlp2020-templates/Sections/Appendix.tex
\section{Packages and Tools}

We use Python 3.7, pytorch 1.7.1, transformers 4.2.2 for the experiments. For tokenization and parsing, we use Spacy 3.0.0 for all the languages. 

\section{Sample statistics}
\label{sec:statistics}

\begin{table}[h]
    \scriptsize
    \begin{tabular}{p{1.6cm}p{0.36cm}p{0.36cm}p{0.36cm}p{0.36cm}p{0.36cm}p{0.36cm}p{0.36cm}}
    \toprule
    \textbf{Perturbations} & \textbf{de} & \textbf{fr} & \textbf{ru} & \textbf{ja} & \textbf{es} & \textbf{it} & \textbf{zh} \\
    \midrule
      TreeMirrorPre   & $3869$  & $3732$ & $3201$ & $1580$ & $7004$ & $3009$ & $155$ \\
      TreeMirrorPost   & $3862$ & $3726$ & $3199$ & $1525$ & $7001$ & $3009$ & $147$ \\
      TreeMirrorIn &  $3862$ & $3726$ & $3199$ & $1525$ & $7001$ & $3009$ & $147$ \\
      VerbAdvSwaps & $944$ & $831$ & $747$ & $1297$ & $1287$ & $615$ & $1649$ \\
      VerbSwaps & $2019$ & $2084$ & $1496$ & $4376$ & $3714$ & $1582$ & $3703$ \\
      NounAdjSwaps & $508$ & $967$ & $631$ & $985$ & $1863$ & $600$ & $469$ \\
      FuncShuffle & $1197$ & $1274$ & $383$ & $7004$ & $2666$ & $666$ & $229$ \\
      NounVerbSwaps & $3777$ & $3624$ & $2821$ & $4798$ & $6664$ & $2687$ & $6746$ \\
      NounVerbMis & $3005$ & $2989$ & $2623$ & $4102$ & $5448$ & $2189$ & $5932$ \\
      ShuffleLastHalf & $3905$ & $4002$ & $3213$ & $4997$ & $7083$ & $3030$ & $7084$ \\
      VerbAtBeginning & $3584$ & $3410$ & $3939$ & $1817$ & $7135$ & $3729$ & $7084$ \\
      RotateAroundRt & $3904$ & $4002$ & $3212$ & $4997$ & $7082$ & $3030$ & $7074$ \\
      WordShuffle & $3905$ & $4002$ & $3213$ & $4997$ & $7083$ & $3030$ & $7084$ \\
      ShuffleFirstHalf & $3905$ & $4002$ & $3213$ & $4997$ & $7083$ & $3030$ & $7084$ \\
      NounSwaps & $3747$ & $2242$ & $2954$ & $4912$ & $5936$ & $1934$ & $6545$ \\
      Reversed & $3904$ & $4002$ & $3212$ & $4997$ & $7082$ & $3030$ & $7074$ \\
      \midrule
      Total & $5k$ & $5k$ & $5k$ & $5k$ & $10k$ & $5k$ & $10k$ \\
    \bottomrule
    \end{tabular}
    \caption{The distribution of samples under different perturbation functions across the different languages. The trend shows that there might be some parts-of-speech that are minority -- Adjective, Adverb -- across the languages. This does not affect the analysis in the paper.}
    \label{tab:sampledist}
\end{table}

\section{$\beta_1$ vs $\beta_2$}
\label{sec:complete-table}

\begin{figure}[h!t]
    \centering  
    \subfigure[EN$\rightarrow$ZH]{
        \includegraphics[width=\columnwidth]{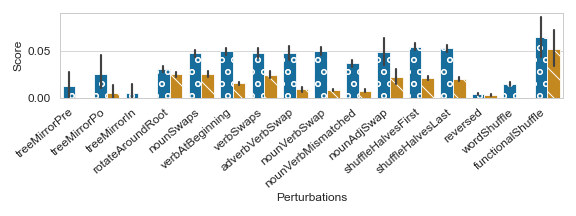}
        }
    \caption{We present the results only from a small set of perturbations to showcase the trend of $\beta_1$ scores being generally higher than $\beta_2$ scores across the different perturbations in different languages.} %This indicates that the translation systems might have a preference to stay robust by ``auto-correcting" the perturbed examples while largely unable to produce faithful translations. The complete result can be found in Appendix.}
    \label{fig:all-perturb-lang-complete}
\end{figure}

\autoref{fig:all-perturb-lang-complete} shows the comparison of the $\beta_1$ and $\beta_2$ scores across the different perturbations on the different translation tasks.

\begin{figure}[h!t]
    \setcounter{subfigure}{0}
    \centering
    \subfigure[EN$\rightarrow$DE]{
        \includegraphics[width=\columnwidth]{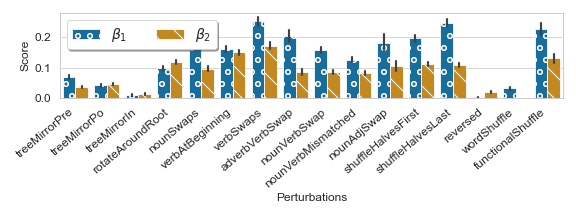}
        }
    \subfigure[EN$\rightarrow$FR]{
        \includegraphics[width=\columnwidth]{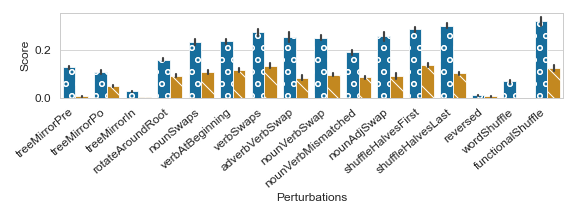}
        }
        
    \subfigure[EN$\rightarrow$RU]{
        \includegraphics[width=\columnwidth]{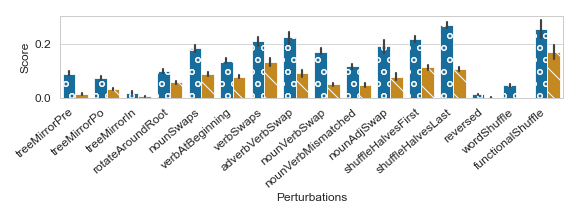}
        }
    \subfigure[EN$\rightarrow$JA]{
        \includegraphics[width=\columnwidth]{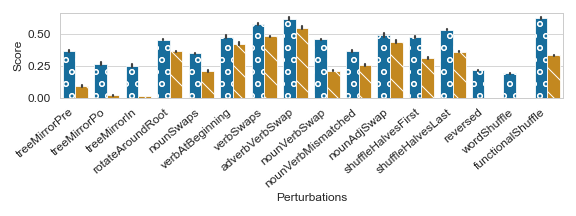}
        }
        
    \subfigure[EN$\rightarrow$IT]{
        \includegraphics[width=\columnwidth]{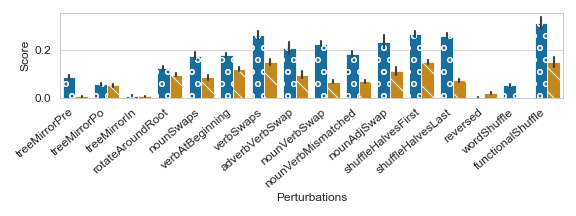}
        }
    \subfigure[EN$\rightarrow$ES]{
        \includegraphics[width=\columnwidth]{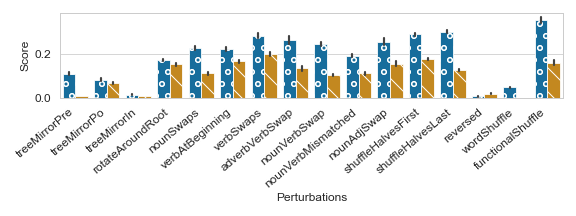}
        }
    \end{figure}
    
\section{$\alpha_e$}
\label{alpha-results}

\begin{figure}[H]
\centering  
\subfigure[$\alpha_e^L$]{
    \includegraphics[width=\columnwidth]{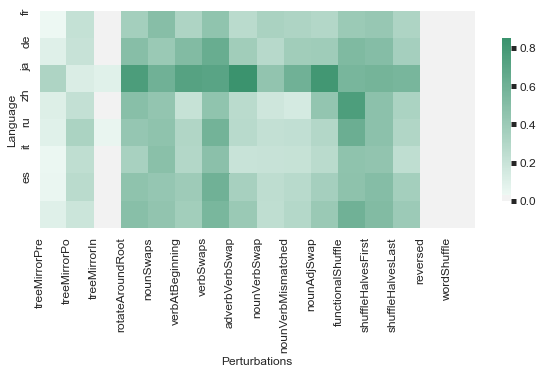}
    }
\caption{The scores of Japanese language shows that the the perturbations on the source text seldom had effects on the sentence and the sentence was close to the original.} %This indicates that the translation systems might have a preference to stay robust by ``auto-correcting" the perturbed examples while largely unable to produce faithful translations. The complete result can be found in Appendix.}
\label{fig:alpha-scores}
\end{figure}

\begin{figure*}[h!t]
    \setcounter{subfigure}{0}
    \centering
    \subfigure[German]{
        \includegraphics[width=0.2\textwidth]{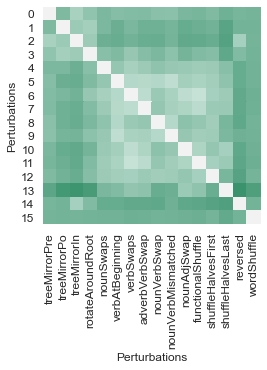}
        }
    \subfigure[French]{
        \includegraphics[width=0.2\textwidth]{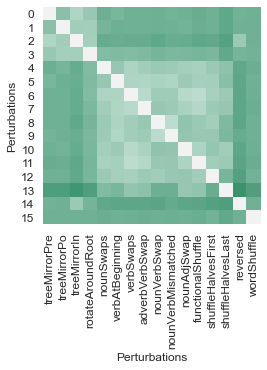}
        }
    \subfigure[Russian]{
        \includegraphics[width=0.2\textwidth]{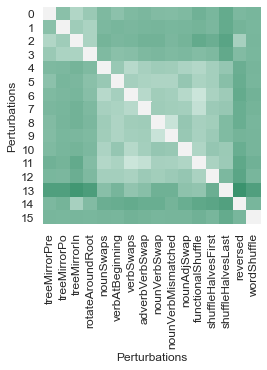}
        }
    \subfigure[Japanese]{
        \includegraphics[width=0.235\textwidth]{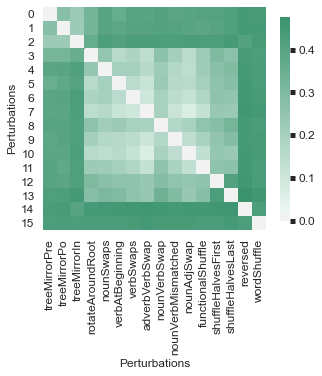}
        }
        
    \subfigure[Italian]{
        \includegraphics[width=0.2\textwidth]{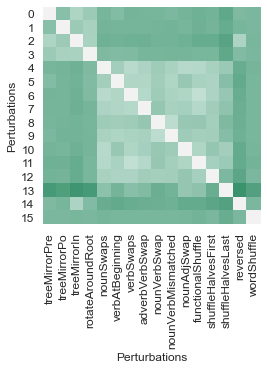}
        }
    \subfigure[Spanish]{
        \includegraphics[width=0.2\textwidth]{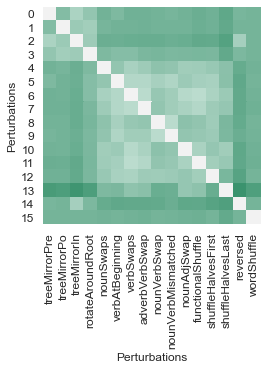}
        }
    \subfigure[Chinese]{
        \includegraphics[width=0.2\textwidth]{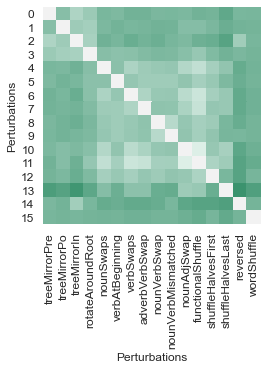}
        }
    \subfigure[English]{
        \includegraphics[width=0.235
        \textwidth]{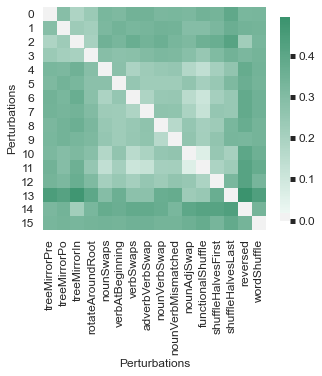}
        }
    % \subfigure[EN-FR]{
    %     \includegraphics{emnlp2020-templates/graphs/fr_bleu-4.png}
    %     % \label{sub_bleu_frames}
    % }
    \caption{The heatmap illustrates average of Levenshtein distances between different perturbations. The map shows interesting patterns that naturally differentiates the dependency tree based, PoS-based and random perturbation categories. It is interesting to observe the pattern being consistent over the different languages.}
    \label{fig:bins-beta-and-alpha-comparison-1}
    
    \centering
    \subfigure[$\beta_1$ vs $\alpha_e$]{
         \includegraphics[width=0.3\textwidth]{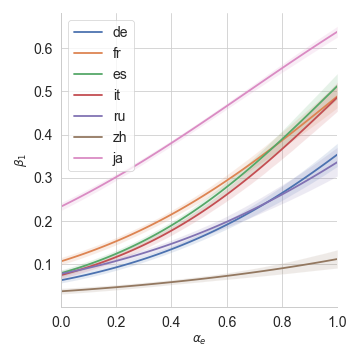}
         }
     \subfigure[$\beta_2$ vs $\alpha_e$]{
         \includegraphics[width=0.3\textwidth]{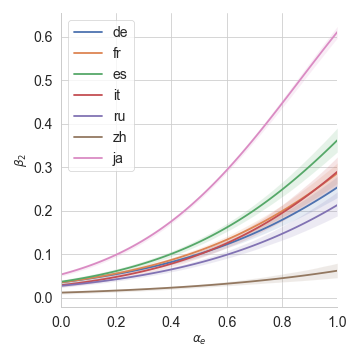}
         }
    \subfigure[$Length$ vs $\alpha_e$]{
         \includegraphics[width=0.3\textwidth]{emnlp2020-templates/graphs/correlation_plots/correlation_Length_and_alpha_e_new.png}
         }
    \caption{The correlation between the $\beta_1$ vs $\alpha_e$ show that the models are staying robust only when the perturbation difficulty is lower ($\alpha_e$ is higher). Despite the perturbations retaining all the words without deletion, staying robust seems hard.}
    \label{fig:other-correlation-plots}
\end{figure*}

\section{Measured MT Performances}

\begin{table}[h!t]
    \centering
    \begin{tabular}{cc}
        \toprule
         {\bf Language} & {\bf BLEU-4} \\
        \midrule
         German &  $0.40 \pm \num{7.77E-6}$\\
         Russian & $0.42 \pm \num{1.71E-6}$\\
         French & $0.43 \pm \num{2.86E-6}$\\
         Japanese & $0.40 \pm \num{1.01E-6}$\\
         Italian & $0.43 \pm \num{2.08E-6}$\\
         Spanish & $0.44 \pm \num{2.64E-6}$\\
         Chinese & $0.32 \pm \num{1.29E-6}$\\
         \bottomrule
    \end{tabular}
    \caption{The performances of the MT systems used in the experiments and their measured performances in BLEU-4 ($\beta$).}
    \label{tab:recorded-mt-performances}
\end{table}

\section{$\beta_1$ $>$ $\beta$}

In some corner cases, we observed the $\beta_1$ to be greater than $\beta$. This suggests that the model, at least in those cases, opts an unnatural understanding of the syntax for the translation.

\begin{table*}[h!t]
    \centering
    \small
    \begin{tabular}{cp{5cm}p{5cm}ccl}
    \toprule
        {\bf Task} & $g_e$ & $g_e^-$ & $\beta$ & $\beta_1$ & $\Psi$ \\
    \hline
        \multirow{3}{*}{de}& Did you bring a hair dryer? & a hair dryer Did you bring ? & $0.00$ & $0.54$ & treeMirrorPo \\ 
        & It's a river that has never been explored. &	It 's a river that has explored been never .& $0.42$ & $0.59$ & nounVerbSwap \\
        & I may go to Boston next month. &	go may I to Boston next month . & $0.37$ & $0.52$ & nounVerbMis\\
        \hline
        \multirow{3}{*}{fr}  & Yes, my name is Karen Smith.	&  Karen Smith Yes , my name is . & $0.00$ & $0.61$ & treeMirrorPo \\
        & Why didn't you call me last night? &	did you n't Why call me last night ? & $0.50$ & $1.00$ & shuffleFirstHalf \\
        & Our fridge doesn't work anymore. & does Our fridge n't work anymore . & $0.00$ & $0.54$ & nounVerbSwap \\
        \hline
        \multirow{3}{*}{es} & Have you ever been on TV? &	been Have you ever on TV ? & $0.34$ & $0.62$ & verbAtBeginning \\
        & I'm looking forward to your coming to Japan. &	I coming looking forward to your 'm to Japan . & $0.45$ & $0.51$ & verbSwaps \\
        & We left him some cake. &	We some left cake him . & $0.0$ & $0.54$ & wordShuffle  \\
        \hline
        \multirow{3}{*}{it} & Have you tried online dating? &	you Have tried online dating ? & $0.45$ & $0.76$ & nounVerbSwap \\
        & What did you do this morning? &	What this do you morning did ? & $0.00$ & $1.00$ & wordShuffle \\
        & She was able to read the book. &	read She was able to the book . & $0.35$ & $0.65$ & verbAtBeginning \\
        \hline
        \multirow{3}{*}{ru} & Tom knew that I was lonely. &	Tom knew that lonely was I . & $0.43$ & $0.64$ & nounAdjSwap \\
        & He said he would come tomorrow. & come he said would He tomorrow . & $0.47$ & $1.00$ & nounVerbMis \\
        & You can stay if I want to. & You can stay if to want I. & $0.45$ & $0.54$ & shuffleHalvesLast \\
        \hline
        \multirow{3}{*}{ja} & Joseph said to them, "It is like I told you, saying, 'You are spies!' &	Joseph saying to them , " It are like I said you , is , ' You told spies ! ' & $0.48$ & $1.00$ & verbSwaps \\
        & Don't be overcome by evil, but overcome evil with good. &	Do n't overcome overcome by evil , but be evil with good . & $0.42$ & $1.00$ & verbSwaps \\
        & How amiable are thy tabernacles, O LORD of hosts! & hosts of LORD O , tabernacles thy are amiable How ! & $0.42$ & $0.65$ & reversed \\
        \hline
        \multirow{3}{*}{zh} & He has completely lost all sense of duty. &	He has lost completely all sense of duty . & $0.45$ & $0.54$ & verbAdverbSwap \\
        & We have a white cat. & We have a cat white . & $0.35$ & $0.84$ & nounAdjSwap\\
        & The main question is how does Tom feel. &	The main question does how is Tom feel . & $0.47$ & $0.61$ & verbSwaps\\
        \bottomrule
    \end{tabular}
    \caption{Samples from across different languages and perturbations where the models translated better when the source sentence was perturbed. Although such flips made only a small fraction, we observed the unnaturalness understanding of the syntactic structure in translation task. This is similar to the observations made by \citet{sinha2020unnatural}.}
    \label{tab:flips-NMT}
\end{table*}